\documentclass[10pt,twocolumn,letterpaper]{article}

\usepackage{cvpr}              

\usepackage{graphicx}
\usepackage{amsmath}
\usepackage{amssymb}
\usepackage{booktabs}
\usepackage{amssymb}
\usepackage{xcolor}
\usepackage{bm}
\usepackage{epigraph}
\usepackage{overpic}
\usepackage{array}

\usepackage{multirow}

\usepackage{enumitem}
\usepackage{pifont}
\usepackage{comment}

\usepackage{mathtools} 
\usepackage{cuted} 

\usepackage{epigraph}

\definecolor{cvprblue}{rgb}{0.21,0.49,0.74}
\usepackage[pagebackref,breaklinks,colorlinks,citecolor=cvprblue]{hyperref}

\usepackage[capitalize]{cleveref}
\crefname{section}{Sec.}{Secs.}
\crefname{section}{Section}{Sections}
\crefname{table}{Table}{Tables}
\crefname{table}{Tab.}{Tabs.}

\def\paperID{4375} 
\def\confName{CVPR}
\def\confYear{2024}

\begin{document}
\newcommand{\mx}{\mathbf{x}}
\newcommand{\my}{\mathbf{y}}
\newcommand{\mz}{\mathbf{z}}
\newcommand{\ms}{\mathbf{s}}
\newcommand{\mI}{\mathbf{I}}
\newcommand{\cD}{\mathcal{D}}
\newcommand{\cM}{\mathcal{M}}
\newcommand{\cN}{\mathcal{N}}
\newcommand{\cG}{\mathcal{G}}

\newcommand{\mepsilon}{\bm{\epsilon}}
\newcommand{\rmbin}{\mathrm{bin}}

\newcommand{\mxh}{\hat{\mx}}
\newcommand{\myh}{\hat{\my}}
\newcommand{\tK}{\tilde{K}}

\newcommand{\mxx}[1]{\mx_{#1}}
\newcommand{\myy}[1]{\my_{#1}}
\newcommand{\mxi}{\mxx{i}}
\newcommand{\myi}{\myy{i}}
\newcommand{\mxhi}{\mxh_i}
\newcommand{\myhi}{\myh_i}
\newcommand{\mxt}{\mxx{t}}

\newcommand{\bbR}{\mathbb{R}}
\newcommand{\bbN}{\mathbb{N}}
\newcommand{\pz}{\phantom{0}}

\title{SatSynth: Augmenting Image-Mask Pairs through Diffusion Models \\for Aerial Semantic Segmentation}
\author{Aysim Toker$^1$\qquad Marvin Eisenberger$^1$\qquad Daniel Cremers$^1$\qquad Laura Leal-Taixé$^2$\\[5pt]
$^1$Technical University of Munich\qquad $^2$NVIDIA
}
\maketitle
\begin{strip}
    \centerline{
  \footnotesize
  \begin{tabular}{c}
    \setlength{\tabcolsep}{0pt}
    \begin{overpic}
        [width=\textwidth]{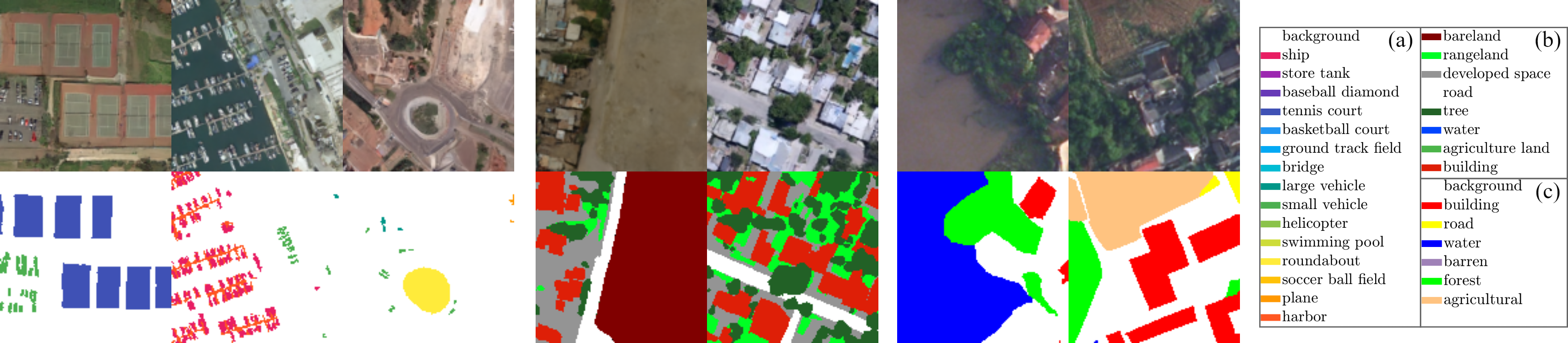}
        \put(10.5,23){\normalsize(a) iSAID~\cite{waqas2019isaid}}
        \put(35.5,23){\normalsize(b) OpenEarthMap~\cite{xia2023openearthmap}}
        \put(61,23){\normalsize(c) LoveDA~\cite{wang2021loveda}}
        \put(-3,13){\rotatebox{90}{\normalsize Image $\mx'$}}
        \put(-3,2.75){\rotatebox{90}{\normalsize Mask $\my'$}}
    \end{overpic}
    \end{tabular}
  }
  \vspace{-0.15cm}
\captionof{figure}{We leverage generative image diffusion to synthesize novel training data instances $(\mx',\my')\sim p(\mx,\my)$ for a given labeled earth observation dataset~\cite{waqas2019isaid,xia2023openearthmap,wang2021loveda}. In our experiments, we demonstrate that integrating such synthetic pairs as training data for downstream semantic segmentation yields significant quantitative improvements.}
\vspace{-0.15cm}
\label{fig:teaser}
\end{strip}

\begin{abstract}
In recent years, semantic segmentation has become a pivotal tool in processing and interpreting satellite imagery. Yet, a prevalent limitation of supervised learning techniques remains the need for extensive manual annotations by experts. In this work, we explore the potential of generative image diffusion to address the scarcity of annotated data in earth observation tasks. The main idea is to learn the joint data manifold of images and labels, leveraging recent advancements in denoising diffusion probabilistic models. To the best of our knowledge, we are the first to generate both images and corresponding masks for satellite segmentation. We find that the obtained pairs not only display high quality in fine-scale features but also ensure a wide sampling diversity. Both aspects are crucial for earth observation data, where semantic classes can vary severely in scale and occurrence frequency. We employ the novel data instances for downstream segmentation, as a form of data augmentation. In our experiments, we provide comparisons to prior works based on discriminative diffusion models or GANs. We demonstrate that integrating generated samples yields significant quantitative improvements for satellite semantic segmentation -- both compared to baselines and when training only on the original data.
\end{abstract}
 
\vspace{-0.2cm}
\section{Introduction}
\label{sec:intro}

Satellite imagery is a powerful tool to monitor the earth's surface, both in terms of specific events and global trends in land use. This has direct implications for humanitarian challenges such as disaster response, food security, and quantifying the impact of climate change. The United Nations summarizes a number of landmark objectives in its Sustainable Development Goals (SDGs)\footnote{https://www.un.org/sustainabledevelopment/sustainable-development-goals/}, in which they include general long-term goals such as the accessibility of clean water (SDG-6), reducing carbon emissions (SDG-13), or maintaining forests and combating desertification (SDG-15). In this context, satellite data can help provide crucial insights for monitoring progress, facilitating more targeted interventions. At the beginning of August $2023$, the extent of Antarctic sea ice was observed to be $2.4\text{ million km}^2$ less than the mean value from previous records ($1979$ to $2022$), an area larger than Greenland. Monitoring this alarming trend, now commonly referred to as a five-sigma event among experts, was largely made possible through the wide-spread use of satellite observations.

While raw satellite data is readily available from various sources, obtaining corresponding semantic labels is challenging and costly due to the need for extensive manual annotation. Such labels are imperative for numerous applications, since they enable us to reason about the semantic content of satellite scenes, particularly through supervised learning. For instance, the relative extent of Antarctic sea ice is directly reflected in respective land-cover annotations. A common solution is to leverage data augmentation to increase the sample diversity and make optimal use of existing labels. However, conventional image augmentation techniques designed for object-centric data -- such as flipping, rotating, and rescaling -- are often insufficient to emulate the large sample diversity of satellite imagery. Individual scenes typically contain a multitude of different object instances and land-cover categories. Semantic correlations within these images are mostly local. This vast diversity implies that most existing datasets are sparse, covering only a fraction of potential earth observation scenes. 

In this work, we advocate for enhancing semantic segmentation of satellite data by harnessing recent advances in generative diffusion models~\cite{ho2020denoising}.
Such models approximate the distribution of an existing dataset, to produce novel samples. 
We demonstrate that this can be leveraged to train a model that imitates the joint distribution of images and semantic segmentation labels in a given satellite dataset. Sampling from this distribution effectively enables us to generate additional training data as a form of data augmentation, see~\cref{fig:teaser} for several such sample pairs. The enhanced training set can then be utilized for downstream semantic segmentation. In a broader context, our work serves as a study of the potential of image diffusion for data synthesis when annotations are scarce and costly.

\paragraph{Contributions.}
\begin{enumerate}
    \item For a given earth observation dataset, we propose to learn the joint data distribution $p(\mx,\my)$ of images $\mx$ and labels in bit space $\my$ via a diffusion model $\cG$.
    \item We employ $\cG$ to generate novel training data instances as a form of data augmentation to enhance downstream semantic segmentation.
    \item We demonstrate that integrating the synthesized pairs yields significant quantitative improvements on three satellite benchmarks~\cite{waqas2019isaid,wang2021loveda,xia2023openearthmap}.
\end{enumerate}
\section{Related work}
\label{sec:related_work}

\paragraph{Denoising diffusion models.} In recent times, diffusion models have emerged as a central technique for image generation. Their main advantage is the ability to synthesize high-quality samples, on par with GANs, while being less prone to suffer from mode collapse~\cite{ho2020denoising,sohl2015deep,song2020denoising,dhariwal2021diffusion}. Besides unconditional generation of novel data instances, common applications of image diffusion models include inpainting~\cite{nichol2021glide,lugmayr2022repaint}, super-resolution~\cite{ho2022cascaded}, style transfer~\cite{zhang2023inversion}, depth prediction~\cite{saxena2023monocular}, and general x-to-image tasks~\cite{rombach2022high,ramesh2022hierarchical,saharia2022photorealistic}.
In the literature, there is a specific interest in generating images conditioned on textual input. Some well-known approaches include: Stable diffusion~\cite{rombach2022high}, DALLE-2~\cite{ramesh2022hierarchical}, Imagen~\cite{saharia2022photorealistic}, Imagen Video~\cite{ho2022imagen}, and GLIDE~\cite{nichol2021glide}.

\paragraph{Semantic segmentation.} Segmenting images, \ie, assigning semantic labels to each pixel in 2D space, is a key challenge in computer vision.
Supervised learning has emerged as the central paradigm for this task, as evidenced by numerous  approaches~\cite{badrinarayanan2017tpami,chen2017deeplab,chen2017rethinking,chen2018eccv,ronneberger2015u,zhao2017pyramid,bertasius2016semantic,takikawa2019gated,lin2017refinenet} and benchmark datasets~\cite{cordts2016cityscapes,mottaghi2014role,zhou2017scene,lin2014microsoft}.
For a comprehensive review, we refer the reader to a recent survey~\cite{csurka2022semantic}.

Compared to mainstream computer vision datasets, satellite imagery is subject to unique challenges. Due to the limited resolution of individual objects, distinct classes often exhibit a high visual similarity~\cite{zheng2020foreground,yang2022sparse}. Moreover, scale variations between different object categories can result in class imbalances. The relevant foreground classes are often dominated by a much larger background class. For instance, the ratio of foreground pixels is $29.75\%$ in VOC2012~\cite{everingham2015pascal}, whereas it is only $2.85\%$ in the iSAID satellite dataset~\cite{waqas2019isaid,zheng2020foreground}. Common solutions to these challenges involve specific architectures that emphasize salient foreground object features and increase robustness to background noise~\cite{zheng2020foreground,li2021pointflow,yang2022sparse}. 

Despite the significant progress, supervised learning approaches still require dense ground-truth labels, which are both costly and difficult to obtain. To decrease the demand for annotated data, several works~\cite{ayush2021geography,manas2021seasonal,mall2023change} explore self-supervised learning. Most such approaches harness some of the unique characteristics inherent to satellite imagery, such as time-series observations~\cite{toker2022dynamicearthnet} or geolocation meta-data~\cite{ayush2021geography}. Similarly, SatMAE~\cite{cong2022satmae} proposes large-scale pre-training for spectral and temporal satellite images using masked autoencoders~\cite{he2022masked}.
Rather than leveraging self-supervision for representation learning and fine-tuning for segmentation, we instead propose to synthesize labeled data points directly via a generative image diffusion model.

\paragraph{Synthesizing training data.} Building on the success of text-to-image diffusion models, a number of works employ them to enhance discriminative tasks. For instance,~\cite{he2022synthetic} explores whether such models are suitable for image recognition in the data-scarce regime. Specifically, they conduct zero-shot and transfer learning experiments, formulating text prompts derived from class label names. In a similar vein,~\cite{bansal2023leaving,azizi2023synthetic} generate text-image pairs, which are then used as additional training data for image classification.

Beyond diffusion models, generating synthetic datasets is a well-established technique to produce vast amounts of labeled data with minimal human input. Several works~\cite{puig2018virtualhome,richter2016playing,gaidon2016virtual,ros2016synthia,zheng2020structured3d} use 3D graphic engines to emulate real-world scenes.

Other approaches~\cite{zhang2021datasetgan,li2022bigdatasetgan} utilize generative adversarial networks (GANs)~\cite{goodfellow2014generative} to synthesize high-quality datasets. Similar to ours, SemGAN~\cite{li2021semantic} fits the training distribution for both images and masks. They then apply test-time optimization for a given query image and extract semantic labels by aligning the learned feature embeddings. In contrast, we propose to directly extract joint pairs of data instances and employ them as training data for downstream semantic segmentation.

\paragraph{Image diffusion for segmentation.}
There are several ways in which diffusion models can improve segmentation tasks. A common strategy is to obtain masks conditioned on input images, adopting the standard image-to-image translation paradigm~\cite{amit2021segdiff}. This approach was subsequently extended to synthesize videos and panoptic masks~\cite{chen2022generalist} for the discrete data representation introduced in AnalogBits~\cite{chen2022analog}. 
Another related line of works leverage pre-trained text-to-image models, such as stable diffusion, with tailored text prompts to address open vocabulary segmentation~\cite{karazija2023diffusion,wu2023diffumask,xu2023open} and object-centric segmentation~\cite{tan2023diffss,nguyen2024dataset}.
Alternatively, some methods leverage diffusion models for self-supervised pre-training. The resulting feature embeddings are subsequently fine-tuned for downstream applications, such as image segmentation~\cite{baranchuk2021label,brempong2022denoising} or binary change detection~\cite{bandara2022ddpm}.
To the best of our knowledge, ours is the first approach that employs diffusion models to jointly generate satellite scenes and the corresponding labels to augment a given training dataset for semantic segmentation. 
\section{Preliminaries}
\label{sec:preliminary}
We provide a brief overview of denoising diffusion probabilistic models (DDPM) as presented in~\cite{ho2020denoising}. The main idea is to devise a generative model capable of synthesizing images by reversing a stochastic Gaussian noising process
\begin{equation}
    \mxt:=\sqrt{1-\beta_t}\mxx{t-1}+\sqrt{\beta_t}\mepsilon\text{, where }\mepsilon\sim\cN(0,\mI).
\end{equation} 
Here, the hyperparameter $\beta_t>0$ specifies the noise variance at each timestep.
The model's forward process follows a predefined schedule of progressive noising steps $\mxx{0}\to\mxx{1}\to\dots\to\mxx{T}$. For sufficiently large $T$, we effectively obtain a Gaussian random sample $\lim_{T\to\infty}\mxx{T}\sim\cN\bigl(0,\mI\bigr)$. This noising process is subsequently reversed $\mxx{T}\to\dots\to\mxx{0}$ by utilizing a U-Net backbone~\cite{ronneberger2015u} to predict the noise vectors from the noisy samples $\mxx{t}$. Through appropriate reparameterization, it effectively learns a mapping $\mxx{t}\mapsto\mxx{t-1}$.
Starting from a random sample $\mxx{T}$, this results in an iterative inverse process 
\begin{equation}\label{eq:ddpmmapping}
    \cG:\bbR^{L}\to\bbR^{H \times W \times C}.
\end{equation}
This stochastic generative model $\cG$ maps from a predefined noise distribution $\mz\sim\mu:=\cN(0,\mI_L)$ to synthesized images $\mx:=\cG(\mz)$ that follow the input training distribution. In the most general setting, we typically have $L=THWC$.  
For further technical details and in-depth explanations of DDPM, we refer the reader to the original publication~\cite{ho2020denoising}. 
\section{Method}
\label{sec:method}
\subsection{Problem statement}\label{subsec:problemstatement}
In this work, we address the task of semantic segmentation for earth observation data. Specifically, we consider a dataset 
\begin{multline}
    \cD:=\bigl\{(\mxi,\myi)\bigr|\mxi\in\bbR^{H \times W \times 3}, \\\myi\in\{0,\dots,K-1\}^{H \times W}, 1\leq i\leq N\},
\end{multline}
of $N$ satellite images $\mxi$ and corresponding semantic maps $\myi$ with a spatial size of $H \times W$, and $K$ distinct semantic classes.

We further assume that $\cD$ is sampled from an underlying latent data manifold $\cM$, which we can only access indirectly through the given instances $(\mxi,\myi)\sim\cM$. The goal of semantic segmentation approaches is to devise a discriminative model for the inference task specified by $p(\my|\mx)$. Standard supervised learning approaches estimate this conditional probability by training a model on the discrete set of samples $\cD\subset\cM$. This approach, however, requires that $\cD$ yields an adequate coverage, \ie, is sufficiently large and diverse. This poses a significant challenge for earth observation, where labeled data is often limited. 

\begin{figure}
    \centering
    \begin{overpic}
        [width=1.0\linewidth]{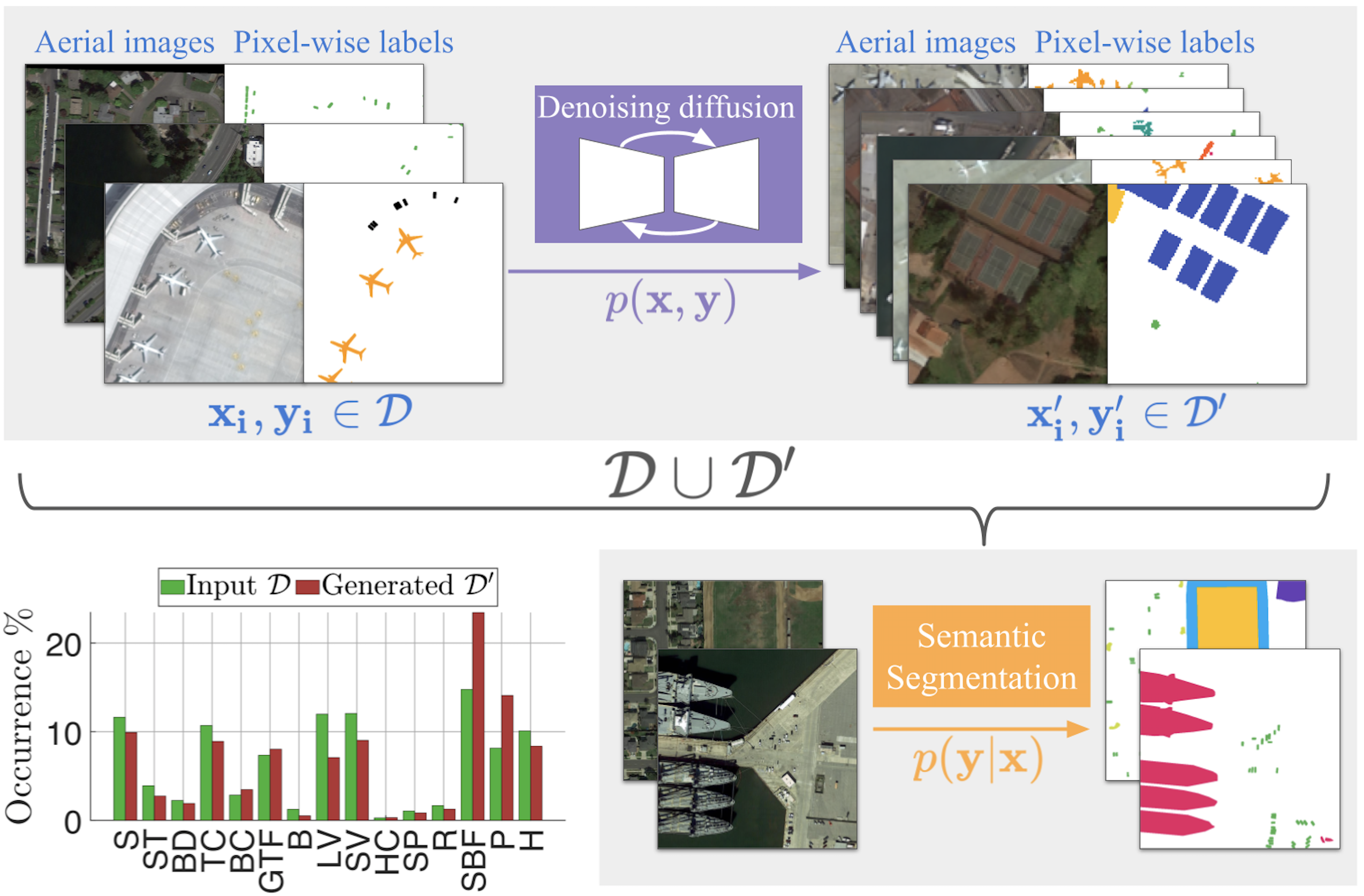}
        \put(1,34){\small (a)}
        \put(44,1){\small (b)}
        \put(1,1){\small (c)}
    \end{overpic}
    \caption{\textbf{Approach overview.} (a) We train a generative image diffusion model $\cG$ on the joint data instances $(\mxi,\myi)\in\cD$ of images $\mxi$ and corresponding labels $\myi$. We then employ $\cG$ to generate a dataset $\cD'$ of novel training samples $(\mxi',\myi')$. (b) Both the real $\cD$ and generated $\cD'$ pairs are integrated and leveraged for the downstream semantic segmentation task. (c) Moreover, we compare the resulting distributions of foreground classes, highlighting that the set of generated labels in $\cD'$ closely matches the original distribution $\cD$. For a legend of label acronyms, refer to~\cref{fig:teaser} (a). 
    }
    \label{fig:overview_fig}
\end{figure}

\paragraph{Motivation.} In our approach, we circumvent this classical data bottleneck by leveraging recent advances in generative modeling. Instead of predicting $p(\my|\mx)$ directly, we first approximate the joint distribution $p(\mx,\my)$ with an unconditional image diffusion model. The resulting network $\cG$ subsequently enables us to generate novel samples $(\mxi',\myi')\sim\cM$.
We then train a segmentation model on the joint dataset $\cD\cup\cD'$ comprising both synthetic $\cD':=\bigl\{(\mx_1',\my_1'),\dots,(\mx_{N'}',\my_{N'}')\bigr\}$ and real $\cD$ samples. For a graphical representation of our approach, refer to~\cref{fig:overview_fig}.

\subsection{Discrete labels in bit-space}\label{subsec:discretelabelsbit}

Most conventional generative models for image data focus on synthesizing instances from the source image distribution $p(\mx)$. Instead, we propose to generate new training instances according to the joint probability $p(\mx,\my)$ of images $\mx$ and corresponding labels $\my$. Extending standard generative models from $\mx$ to $(\mx,\my)$ is not straightforward, since each pixel in $\my$ is associated with a discrete label $\{0,\dots,K-1\}$, as opposed to continuous values in $\bbR$. We circumvent this issue by modeling the discrete values $\my$ in terms of their binary code, specified as 
\begin{equation}\label{eq:analogbits}
    \rmbin:\{0,\dots,K-1\}\to\{0,1\}^{\lceil \log_2{K} \rceil }.
\end{equation}
Compared to a standard one-hot encoding, this was shown to yield an improved stability for generative models in prior work~\cite{chen2022analog}. It also leads to only a small (logarithmic) increase in dimensionality when considering a high number of classes $K$, see~\cref{tab:ablation_study} for a comparison. In practice, we normalize the RGB values of the input images $\mxi$ to the same range $[0,1]$ for compatibility with the binary values $\rmbin(\myi)$. This facilitates synthesizing both quantities jointly, which we describe in the next section.

\subsection{Synthesizing satellite segmentation data}\label{subsec:approach}

The transformation in~\cref{eq:analogbits} maps the discrete labels to the domain of binary values $\rmbin(\myi)\in\{0,1\}^{H \times W \times \lceil \log_2{K} \rceil}$. This allows us to simply concatenate them with the (normalized) RGB values as additional input channels. In order to synthesize pairs of novel data points $\mxi'$ and $\myi'$, we follow a simple strategy:
\begin{enumerate}
    \item Train a generative model $\cG$ on all joint data instances $\bigl(\mxi,\rmbin(\myi)\bigr)\in[0,1]^{H \times W \times \left(3+\lceil \log_2{K} \rceil\right)}$ in $\cD$. 
    \item Generate novel, synthetic samples $\bigl(\mxhi,\myhi\bigr)\sim\cG$.
    \item Threshold and transform $(\mxi',\myi'):=(\mxhi,\rmbin^{-1}(\myhi))$.
    \item Employ $\cD':=\bigl\{(\mxi',\myi')| 1\leq i\leq N'\}\bigr\}$ for downstream tasks, \eg, train a segmentation model.
\end{enumerate}
While other choices are possible, in this work we leverage recent advances in state-of-the-art diffusion models $\cG$. The exact architecture and training schedule is based on DDPM~\cite{ho2020denoising}, see~\cref{sec:preliminary} for a brief overview.

\subsection{Image super-resolution}\label{subsec:superresolution}
In the context of diffusion models, many architectures that operate in image-space focus on relatively coarse resolutions $H=W\leq128$. However, for many earth observation tasks, a sufficient level of detail is crucial. Although in theory, existing models can be trained for higher resolutions, there are significant practical limitations~\cite{rombach2022high}. For once, the training cost of such models, which are already computationally intensive, increases further. Additionally, it often leads to unstable training behaviour and inferior samples, see~\cref{fig:superres_exp} in the appendix for a comparative analysis.

Instead of training our model for higher resolutions directly, we leverage recent advances in image super-resolution~\cite{ho2022cascaded}.
Rather than introducing a new architecture, we simply employ a conditional variant of the architecture detailed in~\cref{subsec:discretelabelsbit}. 
Specifically, we train a DDPM image-to-image translation model that jointly generates images and labels in bit space, while being conditioned on the corresponding low-resolution samples:
\begin{equation}\label{eq:superres}
    \cG_\mathrm{SR}:\bbR^{L}\times\bbR^{H \times W \times C}\to\bbR^{2H \times 2W \times C}.
\end{equation}
The first input is a random noise vector $\mz\in\bbR^{L}$ analogous to~\cref{eq:ddpmmapping}, whereas the second input is the conditional low-resolution image-mask pair with a channel size of $C=3+\lceil \log_2{K} \rceil$. At each timestep, the current prediction of the denoising U-Net is concatenated with the low-resolution pair. During test time, we generate higher-resolution samples in two consecutive steps: $(\mxhi,\myhi):=\cG_\mathrm{SR}(\mz_1, \cG(\mz_0))$. We provide visualizations of several obtained super-resolution pairs in~\cref{fig:highres}.

\subsection{Implementation details}\label{subsec:implementationdetails}

\paragraph{Diffusion.}
As described above, we employ DDPM~\cite{ho2020denoising} to learn the joint training distribution $\cD$. Specifically, we apply $T=1000$ consecutive denoising steps with a linear noise schedule $\beta_t\in\{1e-4,\dots,2e-2\}$. We consistently generate images with a spatial size of $128\times128$ and optionally upsample them to $256\times256$ as specified in~\cref{subsec:superresolution}.

In~\cref{subsec:approach}, we denote the inverse binary transformation with a slight abuse of notation as $\rmbin^{-1}$. In practice, this involves a combination of thresholding the continuous values $\myhi$, as well as mapping the resulting binary values to the original index domain $\{0,\dots,K-1\}$. Since $K$ is not necessarily a power of $2$, we use a simple nearest-neighbor assignment of $\myhi$ to $\bigl\{\rmbin(0),\dots,\rmbin(K-1)\bigr\}\subset\{0,1\}^{\lceil \log_2{K} \rceil }$ to ensure bijectivity. 

Prior to querying the denoising U-Net backbone, an additional linear transformation $z\mapsto 2z-1$ is applied to the input pairs for an improved numerical stability. At test time, the generated samples are clipped to the range $[-1,1]$, before the inverse transformation $z\mapsto 0.5(z+1)$ maps the values back to the original interval $[0,1]$.

\begin{table*}
\begin{center}
\begin{subtable}{0.56\textwidth}
\begin{center}
\resizebox{!}{0.97cm}{
\centering
\begin{tabular}{lccccccccc}
\toprule[0.1em]
&\multicolumn{3}{c}{\emph{iSAID}} & \multicolumn{3}{c}{\emph{LoveDA}} &  \multicolumn{3}{c}{\emph{OpenEarthMap}} \\ \cline{2-10}
     & \scriptsize FID ($\downarrow$) & \scriptsize  sFID ($\downarrow$) & \scriptsize IS ($\uparrow$) & \scriptsize FID ($\downarrow$) & \scriptsize sFID ($\downarrow$) & \scriptsize IS ($\uparrow$) & \scriptsize FID ($\downarrow$) & \scriptsize sFID ($\downarrow$) & \scriptsize IS ($\uparrow$) \\
    \toprule[0.2em]
    SemGAN~\cite{li2021semantic} & 10.21 & 3.63 & 1.11 & 37.47 & 7.16 & 3.29 & 16.20 & 4.89 & 3.13 \\ 
    DDPM~\cite{ho2020denoising} & 17.50 & 7.07 & 1.03 & 22.70 & 2.10 & 3.44 & 15.35 & 4.87 & \textbf{3.31} \\
    Ours & \textbf{\pz8.66} & \textbf{3.10} & \textbf{1.17} & \textbf{13.87} & \textbf{1.87} & \textbf{3.88} & \textbf{12.09} & \textbf{2.76} & 3.16 \\ 
\bottomrule[0.1em]
\end{tabular}
}
\end{center}
\vspace{-0.2cm}
\caption{Visual sample quality.}
\label{tab:gen_models_comp_a}
\end{subtable}
\hfill
\begin{subtable}{0.43\textwidth}
\begin{center}
\resizebox{!}{0.97cm}{
\begin{tabular}{lcccccc}
\toprule[0.1em]
&\multicolumn{2}{c}{\emph{iSAID}} & \multicolumn{2}{c}{\emph{LoveDA}} &  \multicolumn{2}{c}{\emph{OpenEarthMap}} \\ \cline{2-7}
     & \scriptsize IoU ($\uparrow$) & \scriptsize F1 ($\uparrow$) & \scriptsize IoU ($\uparrow$) & \scriptsize F1 ($\uparrow$) & \scriptsize IoU ($\uparrow$) & \scriptsize F1 ($\uparrow$) \\
    \toprule[0.2em]
    SemGAN~\cite{li2021semantic} & 13.01 & 18.08 & 31.75 & 42.74 & 40.43 & 54.75\\
    SegDiff~\cite{amit2021segdiff} & 41.25 & 54.77 & 36.60 & 49.93 & 51.23 & 65.70\\
    Ours & \textbf{52.13} & \textbf{66.13} & \textbf{48.97} & \textbf{64.83}   & \textbf{62.24} & \textbf{76.10}\\
\bottomrule[0.1em]
\end{tabular}
}
\end{center}
\vspace{-0.2cm}
\caption{Semantic segmentation.}
\label{tab:gen_models_comp_b}
\end{subtable}
\centering
    \caption{\textbf{Comparative analysis of generative models.} We present quantitative comparisons on three distinct satellite benchmarks~\cite{waqas2019isaid,wang2021loveda,xia2023openearthmap}. (a) For once, we assess the visual quality of the synthetic satellite images $\mx_i'\in\bbR^{128 \times 128 \times 3}$ obtained with our approach, compared to images generated by SemGAN~\cite{li2021semantic} and DDPM~\cite{ho2020denoising}. (b) We further quantify the segmentation accuracy of our approach with the FPN segmentation backbone, comparing it to the generative semantic segmentation approaches SemGAN~\cite{li2021semantic} and SegDiff~\cite{amit2021segdiff}. In both tables, ($\downarrow$) indicates lower metric values are better, whereas ($\uparrow$) denotes higher values are better. 
    }
\end{center}

\end{table*}

\begin{figure*}
    \centering
    \includegraphics[width=1.0\textwidth]{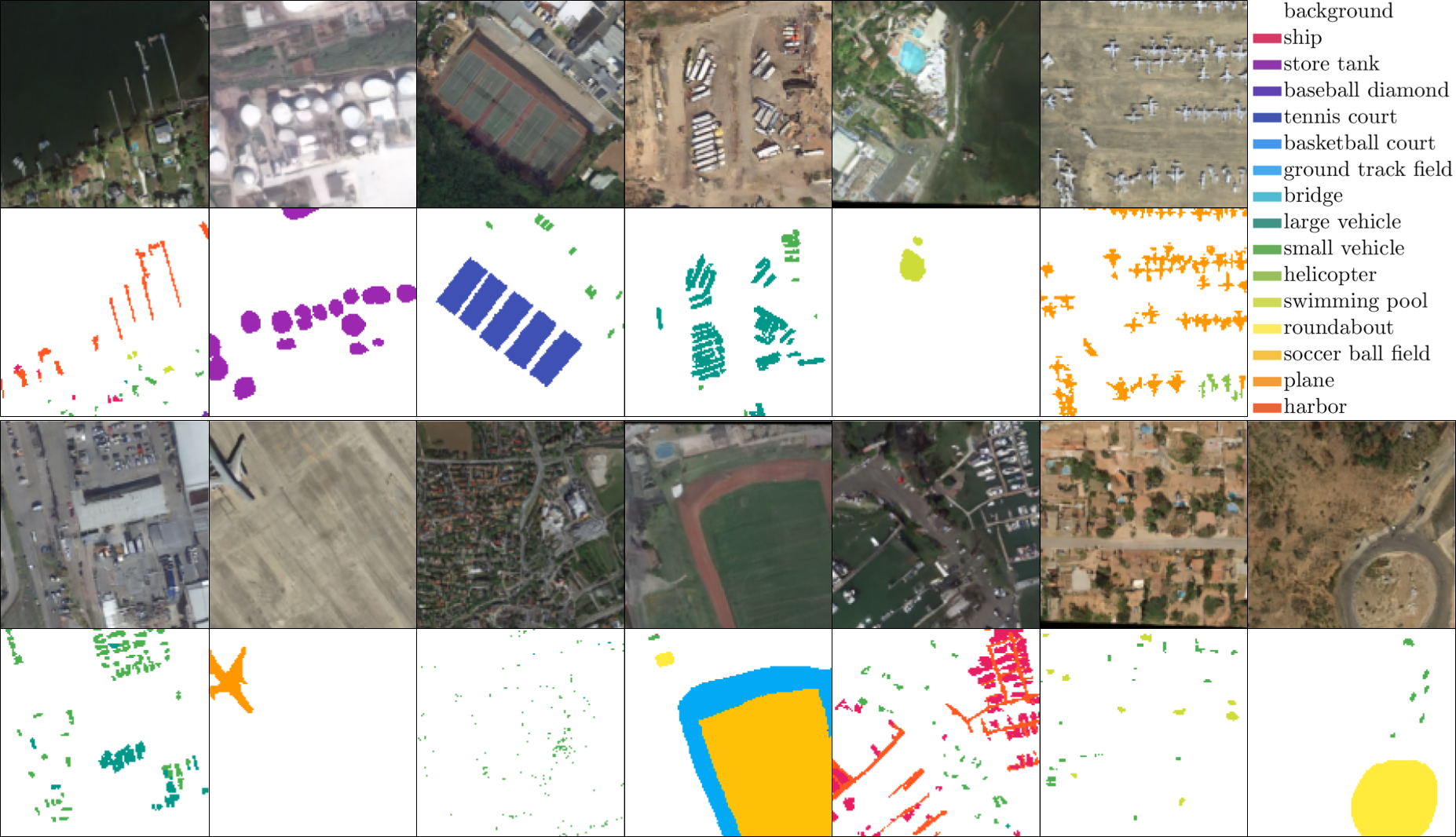}
    \caption{\textbf{Generated samples, iSAID~\cite{waqas2019isaid}.} We visualize several pairs $(\mxi',\myi')$ sampled from the diffusion model $\cG$ detailed in~\cref{subsec:approach}. Color coding for the semantic masks $\myi'$ is indicated by the corresponding palette legend (top right). The generated scenes are of high quality and the semantic layout is coherent -- for instance, soccer ball fields are frequently surrounded by ground track fields (bottom, 4th).
    } 
    \label{fig:syntheticvis_isaid}
\end{figure*}
\paragraph{Segmentation.}
By default, we utilize a standard feature pyramid network (FPN)~\cite{lin2017feature,kirillov2019panoptic} with a ResNet50 backbone to perform multi-class semantic segmentation. The hierarchical design of FPN has proven to be particularly effective in state-of-the-art earth observation approaches~\cite{zheng2020foreground,li2021pointflow}. It allows for predicting accurate masks in the presence of large scale variations and detailed fine-scale structures. 
For a more complete picture, we additionally include a recent vision transformer segmentation model SegFormer~\cite{xie2021segformer} and the state-of-the-art satellite segmentation models PFSegNet~\cite{li2021pointflow} and FarSeg~\cite{zheng2020foreground} for specific settings. 
\section{Experiments}
\label{sec:experiments}

\begin{figure*}
    \vspace{-0.2cm}
    \centering
    \includegraphics[width=0.9\textwidth]{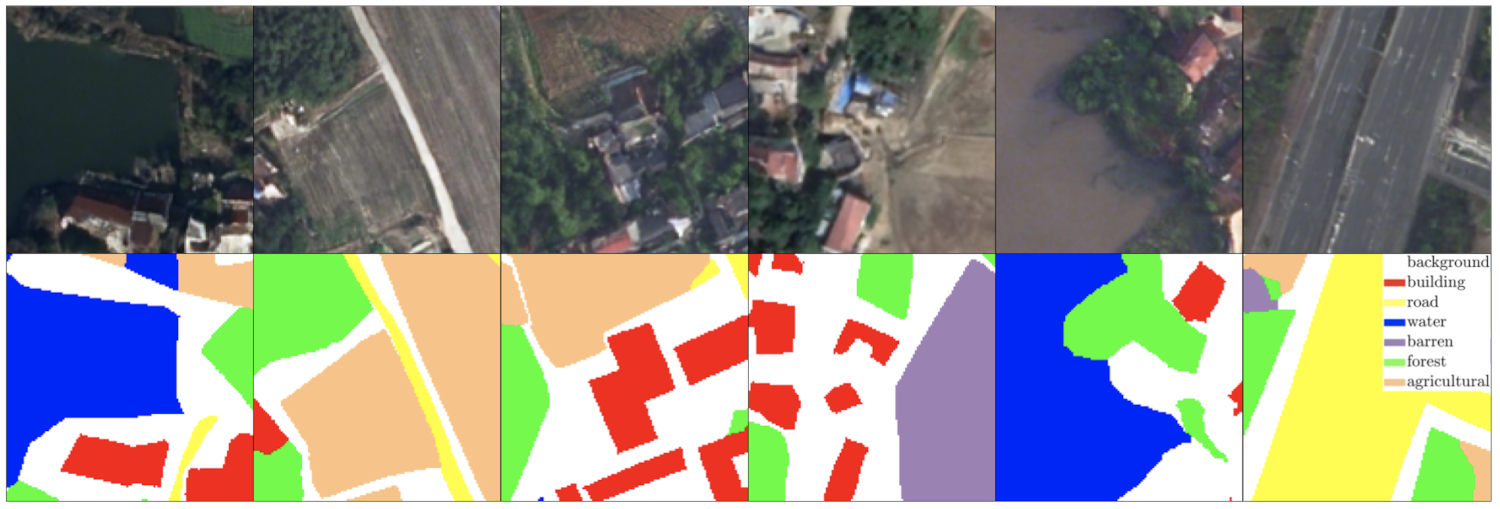}
    \vspace{-0.1cm}
    \caption{\textbf{Generated samples, LoveDA~\cite{wang2021loveda}.} We display pairs $(\mxi',\myi')$ generated by $\cG$ on LoveDA~\cite{wang2021loveda}. The obtained satellite scenes consist of visually plausible images $\mxi'$ and corresponding semantic masks $\myi'$ for general land-cover classes.
    }
    \label{fig:syntheticvis_loveda}
\end{figure*}

We evaluate the impact of synthesized training samples $\cD'$ on the task of semantic segmentation for earth observation data. We provide direct comparisons to prior techniques on generative models, thorough analyses of our generated data, and segmentation results for diverse settings.

\subsection{Datasets}\label{subsec:datasets}
We consider three popular earth observation benchmarks that address object-centric segmentation~\cite{waqas2019isaid} and land-cover classes~\cite{wang2021loveda,xia2023openearthmap}, respectively.

\paragraph{iSAID.} 
The iSAID~\cite{waqas2019isaid} dataset focuses on semantic segmentation of individual object categories such as cars, bridges, or tennis courts. Overall, it contains 2,806 high-resolution satellite images with 655,451 object instances from 15 classes. The original source of the data is the DOTA dataset~\cite{xia2018dota}, while the semantic labels were annotated specifically for iSAID. Since the main goal of the dataset is to segment individual objects, it is prone to large scale variation and class imbalances. 
Certain object categories occupy only a small number of pixels ($\approx 0.3\%$ pixels for `small vehicles') compared to the dominant background class ($>97\%$ pixels).

\paragraph{LoveDA.}
LoveDA~\cite{wang2021loveda} consists of 5,987 high resolution images of both rural and urban scenes, along with 166,768 individual land-cover annotations. A notable challenge are similarities in appearance of distinct categories across different geographical contexts. To increase the number of input images, we extract non-overlapping $256\times 256$ patches from each image, resulting in a total of 40,352 training and 26,704 validation images.

\paragraph{OpenEarthMap.}
The OpenEarthMap~\cite{xia2023openearthmap} dataset integrates high-resolution satellite imagery from several different sources to create a unified benchmark for land-use and land-cover mapping. In particular, they assemble 5,000 images spanning 97 regions across 44 countries from 6 continents. Individual images have a resolution of $1024\times 1024$ with a pixel granularity $\leq 0.5\mathrm{m}$. Similar to LoveDA, we extract $256\times 256$ non-overlapping patches as a pre-processing step. The pixel-wise annotations encompass 8 distinct land-cover categories, namely: bareland, rangeland, developed space, road, tree, water, agriculture land, and buildings. 

\subsection{Comparisons to generative approaches}\label{subsec:expdatasynthesis}
We evaluate our method in two distinct settings. First, we compare our synthesized images $\left\{\mx_1',\dots,\mx_{N'}'\right\}$ to alternative generative approaches in terms of visual sample quality. Second, we assess the impact of our synthesized image-mask pairs on semantic segmentation, contrasting our approach with methods employing generative models for segmentation.

\paragraph{Visual sample quality.}
We compare the sample quality of generated satellite images $\mx_i'\in\bbR^{128 \times 128 \times 3}$ to two baseline approaches~\cite{li2021semantic,ho2020denoising} in~\cref{tab:gen_models_comp_a}. The first baseline, SemGAN~\cite{li2021semantic}, synthesizes image-mask pairs through adversarial training. Additionally, we consider vanilla DDPM~\cite{ho2020denoising}, trained solely on the input images $\left\{\mx_1,\dots,\mx_{N}\right\}$. In each case, we report the Fréchet inception distance (FID)~\cite{heusel2017gans}, spatial FID (sFID)~\cite{nash2021generating}, and the inception score (IS)~\cite{salimans2016improved}, confirming the superior visual quality of our synthesized images.
Specifically, we compute the IS in terms of the segmentation logits of a pretrained ResNet-50 U-Net segmentation model~\cite{ronneberger2015u}. For the FID and sFID scores, we compare the generated feature distribution of the encoder to the distribution of real images $\left\{\mx_1,\dots,\mx_{N}\right\}$. We further provide qualitative samples of generated training pairs on iSAID and LoveDA in~\cref{fig:syntheticvis_isaid} and~\cref{fig:syntheticvis_loveda}, and on OpenEarthMap in~\cref{fig:syntheticvis_openearthmap} in the appendix.

\begin{figure*}
    \centering
    \includegraphics[width=0.95\linewidth]{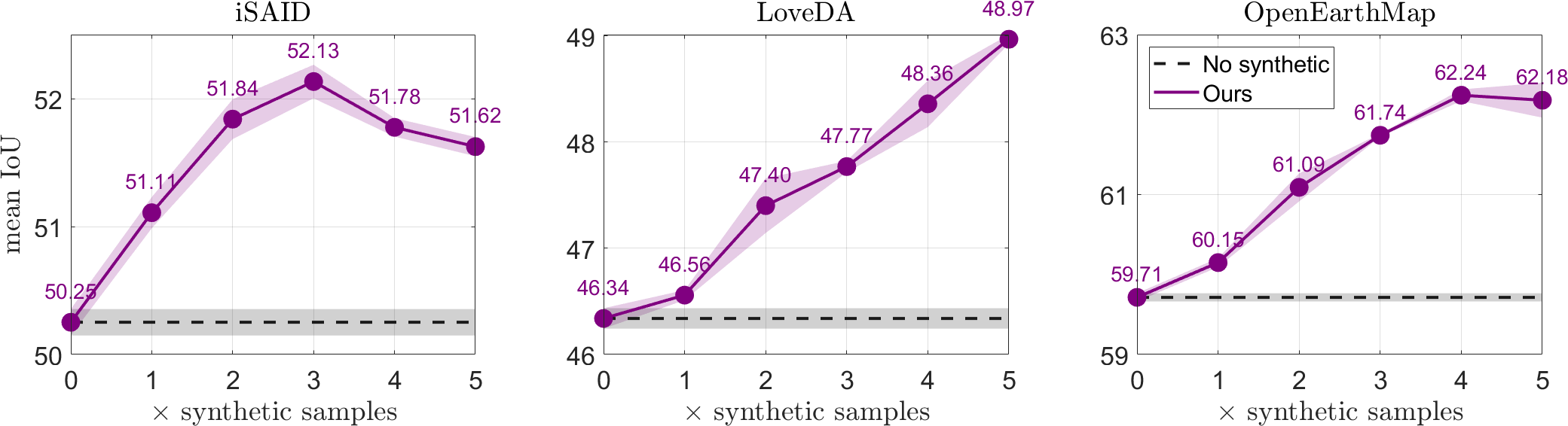}
    \caption{\textbf{Analysis of synthetic data.} 
    We assess the impact of generated samples $\cD\cup\cD'$ on the mIoU segmentation score for iSAID~\cite{waqas2019isaid}, LoveDA~\cite{wang2021loveda}, and OpenEarthMap~\cite{xia2023openearthmap}, with a spatial size of $128\times128$. Different resampling ratios are applied, defined as sampling $R\in\{0,\dots,5\}$ synthetic pairs per original instance, \ie, $|\cD'|=R\cdot|\cD|$ pairs in total. In each case, error bars are provided which denote the standard error (SE). We separately plot the accuracies without synthetic samples $R=0$ (gray dashed lines) for ease of comparison.
    } 
    \label{fig:increasenumberofsynth}
\end{figure*}

\begin{table*}
\begin{center}
\scalebox{0.70}{
\begin{tabular}{lcccccccccc}
\toprule[0.2em]
\emph{Train on}&\multicolumn{2}{c}{PFSegNet~\cite{li2021pointflow}} & \multicolumn{2}{c}{FarSeg~\cite{zheng2020foreground}} &\multicolumn{2}{c} {SegFormer~\cite{xie2021segformer}} &\multicolumn{2}{c}{FPN~\cite{kirillov2019panoptic}} & \multicolumn{2}{c}{PSPNet~\cite{zhao2017pyramid}}\\ \cline{2-11}
     \emph{$256\times256$}&IoU ($\uparrow$)& F1 ($\uparrow$) &IoU ($\uparrow$)& F1 ($\uparrow$) & IoU ($\uparrow$)& F1 ($\uparrow$) & IoU ($\uparrow$)& F1 ($\uparrow$) & IoU ($\uparrow$)& F1 ($\uparrow$)\\
    \toprule[0.2em]
    $\cD$  &60.93$\pm$0.62 & 74.10$\pm$0.55& 62.28$\pm$0.27 &75.16$\pm$0.22 &60.95$\pm$0.61& 74.18$\pm$0.55&59.52$\pm$ 0.19&  72.82$\pm$0.15& 48.95$\pm$2.91 & 63.13$\pm$2.83 \\
    $\cD\cup\cD'$ & \textbf{63.71$\pm$0.21} & \textbf{76.37$\pm0.17$} &\textbf{62.95$\pm$0.25} & \textbf{75.72$\pm$0.28} & \textbf{62.13$\pm$0.25} & \textbf{75.10$\pm$0.21} & \textbf{60.65$\pm$0.38} & \textbf{73.69$\pm$0.33} & \textbf{56.54$\pm$1.32} & \textbf{70.16$\pm$1.17}\\

\bottomrule[0.1em]
\end{tabular}
}
\centering
    \caption{\textbf{Object-centric segmentation}. We demonstrate that integrating our generated training pairs $\cD\cup\cD'$ improves the performance over the original data $\cD$ on the iSAID~\cite{waqas2019isaid} benchmark. To this end, we consider the recent state-of-the-art approaches PFSegNet~\cite{li2021pointflow} and FarSeg~\cite{zheng2020foreground} that specialize on satellite segmentation, as well as three generic segmentation models~\cite{xie2021segformer,kirillov2019panoptic,zhao2017pyramid}.
    In each setting, we utilize super-resolution pairs with a spatial size of $256\times256$ as defined in~\cref{subsec:superresolution}.
    }
    \label{tab:sota_seg}
\end{center}
\end{table*}

\paragraph{Generative segmentation.}

Our proposed approach is based on synthesizing novel training data instances $\cD'$. Further, we integrate the augmented training set $\cD\cup\cD'$ for downstream segmentation. While this approach is straightforward, there exist other potential strategies for employing generative models to predict segmentation masks directly. To provide context, we consider two baseline approaches. SemGAN~\cite{li2021semantic} trains a joint generator and a separate encoder, applying test-time optimization to extract semantic labels conditioned on the latent embedding of a given query image. Conversely, SegDiff~\cite{amit2021segdiff} directly models binary segmentation as an image-to-image regression task utilizing DDPM. Although it is not primarily designed for multi-class scenarios, we can extend it in a straightforward manner by predicting labels in bit space, as detailed in~\cref{eq:analogbits}.

We report the resulting accuracies in~\cref{tab:gen_models_comp_b}. For our approach, we train a standard FPN backbone on the augmented training set $\cD\cup\cD'$. Compared to SemGAN~\cite{li2021semantic} and SegDiff~\cite{amit2021segdiff}, our approach yields significantly more accurate predictions, both in terms of the mIoU and F1 score. A visualization of the predicted semantic maps is provided in~\cref{fig:segmentation_isaid} in the appendix. 

\subsection{Analysis of synthetic training data}\label{subsec:expsyntheticdata}
We assess the impact of additional synthetic samples $\cD'$ on satellite semantic segmentation with an FPN backbone. To this end, we experiment with various resampling ratios $|\cD'|=R\cdot|\cD|$, where $R\in\bbN$.
In principle, we can sample an arbitrary number of pairs $(\mxi',\myi')$ from the diffusion model $\cG$ specified in~\cref{subsec:approach}. However, as we generate more samples, they become increasingly correlated and redundant -- yielding diminishing returns.

We report the resulting mIoU scores on different benchmarks~\cite{waqas2019isaid,wang2021loveda,xia2023openearthmap} for values of $R\in\{0,\dots,5\}$ in~\cref{fig:increasenumberofsynth}. In each scenario, we quantify how training on the joint dataset $\cD\cup\cD'$ compares to the original samples $\cD$. We apply oversampling to ensure a balanced ratio between real and synthetic samples. In each setting, we observe that adding synthetic pairs consistently enhances the performance. This confirms our assertion that integrating generated pairs serves as a form of data augmentation. Moreover, the optimal performance varies on specific benchmarks, for instance, a resampling ratio of $R=3$ is ideal for iSAID.

\begin{figure*}
    \centering
    \includegraphics[width=0.87\textwidth]{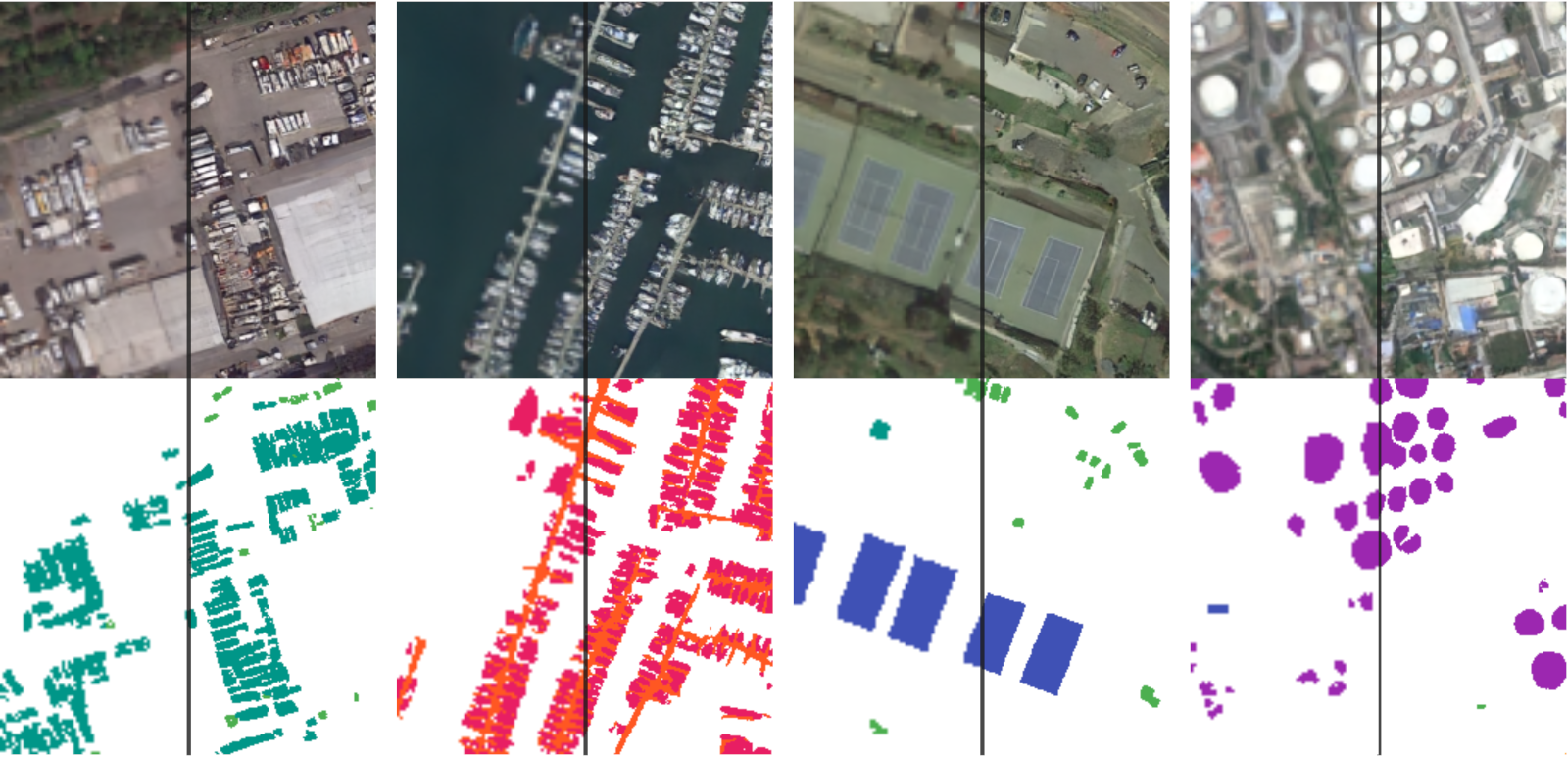}
    \caption{\textbf{Super-resolution, qualitative.} We provide four super-resolution samples obtained on the iSAID dataset. These generated pairs are obtained in two steps, querying $\cG$ to produce low-resolution images with a spatial size of $128\times128$ (left), before upsampling them via $\cG_\mathrm{SR}$ to $256\times256$ (right) as defined in~\cref{eq:superres}. 
    } 
    \label{fig:highres}
\end{figure*}

\subsection{Object-centric segmentation}\label{subsec:expsemanticsegmentation}
We evaluate our approach in terms of object-centric satellite segmentation, see~\cref{tab:sota_seg} for a summary of the resulting accuracies. Specifically, we report the mIoU and F1 scores on the iSAID~\cite{waqas2019isaid} benchmark for several state-of-the-art satellite~\cite{li2021pointflow,zheng2020foreground} and general~\cite{xie2021segformer,kirillov2019panoptic,zhao2017pyramid} segmentation approaches. 
These results demonstrate that training on the augmented dataset $\cD\cup\cD'$ consistently improves the performance over $\cD$. For instance, the mIoU score increases by $7.59\%$ for PSPNet~\cite{zhao2017pyramid}.
Since most considered segmentation methods expect high input resolutions, we leverage our super-resolution model $\cG_\mathrm{SR}$ specified in~\cref{subsec:superresolution} to generate pairs with a spatial size of $256\times256$. While many satellite baselines consider even higher resolutions $>256$ in their original publications~\cite{li2021pointflow,zheng2020foreground}, we found this to be a reasonable trade-off due to the substantial computational demand of high-resolution diffusion models. For qualitative visualizations of different super-resolution image-mask pairs, see~\cref{fig:highres}. 

\subsection{Ablation study}\label{subsec:ablationstudy}
In~\cref{subsec:approach}, we propose to jointly generate images and labels $(\mx,\my)$ for earth observation data via a denoising diffusion model $\cG$. 
We now assess the impact of specific assumptions on the quality of the generated samples, see~\cref{tab:ablation_study} for a summary. Specifically, we compare the binary encoding $\rmbin(\my)$ defined in~\cref{eq:analogbits} to conventional one-hot encoding $\mathrm{OH}(\my)_k:=\mathrm{I}[\my=k]$. The findings in~\cref{tab:ablation_study} indicate that the binary coding yields superior samples. 
We attribute this to the more compact encoding of $\lceil \log_2{K} \rceil$ channels, as opposed to $K$ for one-hot. 
For instance, for the $K=16$ classes in iSAID, the binary codes require only $4$ channels, resulting in well-balanced joint samples $(\mx,\my)$ with a total of $C=7$ channels. 

We further investigate the prediction type targeted by the diffusion U-Net backbone. There exist two prevalent choices, where in each denoising step $\mxt$ the network either predicts the total noise vector $\epsilon$, or the original sample $\mxx{0}$. From both quantities, the noise vector between two timesteps $\mxx{t-1}\to\mxt$ can be derived. On the other hand, we find that, for our purposes, predicting $\epsilon$ leads to a superior segmentation performance -- especially for the higher resampling ratio $R=3$.

\begin{table}
\begin{center}
\resizebox{0.95\linewidth}{!}{
\begin{tabular}{cccc|cccc}
\toprule[0.2em]
 \multicolumn{2}{c}{Label encoding} & \multicolumn{2}{c}{Predict} & \multicolumn{2}{|c}{$R=1$} & \multicolumn{2}{c}{$R=3$} \\ \cline{5-8}
    $\rmbin(\my)$ & $\mathrm{I}[\my=k]$ & $\epsilon$ & $\mxx{0}$ & IoU ($\uparrow$)& F1 ($\uparrow$) & IoU ($\uparrow$)& F1 ($\uparrow$)\\
    \toprule[0.2em]
    & \checkmark & \checkmark & & 47.98 & 62.08 & 49.38 & 63.71 \\
    \checkmark & & & \checkmark & 51.00 & 65.05 & 51.45 & 65.54 \\
    \checkmark & & \checkmark & & \textbf{51.11} & \textbf{65.13} & \textbf{52.13} & \textbf{66.13} \\

\bottomrule[0.1em]
\end{tabular}
}
\centering
    \caption{\textbf{Ablation study.}
    We investigate how two central design choices for the joint data generation, outlined in~\cref{subsec:approach}, relate to the downstream segmentation performance on iSAID. Specifically, we compare two variants of the network $\cG$, predicting the noise component $\epsilon$ or the initial sample $\mxx{0}$, and we contrast the binary label embedding with standard one-hot encoding. These results confirm that both components are crucial (lower row, $\rmbin$+$\epsilon$) for obtaining optimal results. 
    }
    \label{tab:ablation_study}
\end{center}

\end{table}
\section{Conclusion}
Our work serves as a showcase for the potential of image diffusion models for data generation in domains where ground truth labels are scarce and costly. For satellite imagery, specifically, obtaining such labels requires extensive manual annotation by human experts. In a broader context, we anticipate that approaches similar to ours will become increasingly ubiquitous for data synthesis and augmentation tasks in different areas. For instance, an average of 1.5 hours was spent to annotate a single image from Cityscapes, according to the original publication~\cite{cordts2016cityscapes}. 

Given the high visual fidelity of the obtained instances, we plan to explore extensions to other related earth observation tasks in future work. Potential applications include sample data fusion of different observations, image inpainting for cloud removal, or change detection.  

\section*{Acknowledgements}
We acknowledge support by the Munich School for Data Science, the ERC Advanced Grant SIMULACRON and the Munich Center for Machine Learning. 
\label{sec:conclusion}

{
    \small
    \bibliographystyle{ieeenat_fullname}
    \bibliography{ms}
}

\clearpage
\setcounter{page}{1}
\maketitlesupplementary
\appendix

\section{Object-centric segmentation analysis}
Our results of object-centric segmentation on iSAID in~\cref{subsec:expsemanticsegmentation} demonstrate substantial improvements for five separate baseline approaches, including general segmentation models \cite{xie2021segformer,kirillov2019panoptic,zhao2017pyramid} and approaches tailored for satellite imagery~\cite{li2021pointflow,zheng2020foreground}.
We additionally provide a per-class analysis of the results reported in~\cref{tab:sota_seg} of the main paper, summarized in~\cref{tab:allclasses_isaid}. A key insight is that, beyond overall improvements of the average scores, a majority of individual classes benefit from the augmentation. In the extreme case, our approach yields a 15.26\% gain in the IoU score (SBF, PSPNet~\cite{zhao2017pyramid}), whereas the most significant drop in performance is 2.79\% (HC, SegFormer~\cite{xie2021segformer}). For the BC class, the mean and median increase over all baselines is 5.55\% and 4.75\%, respectively.

Remarkably, even for the overall best performing baseline PFSegNet~\cite{li2021pointflow}, our approach still yields significant improvements for all but one classes, with an increased IoU score of up to 8.98\% (basketball court).
We conclude that the observed improvements of leveraging our synthesized data are homogeneous and consistent throughout all considered settings and for most individual classes.

\begin{figure}
    \centering
    \includegraphics[width=\linewidth]{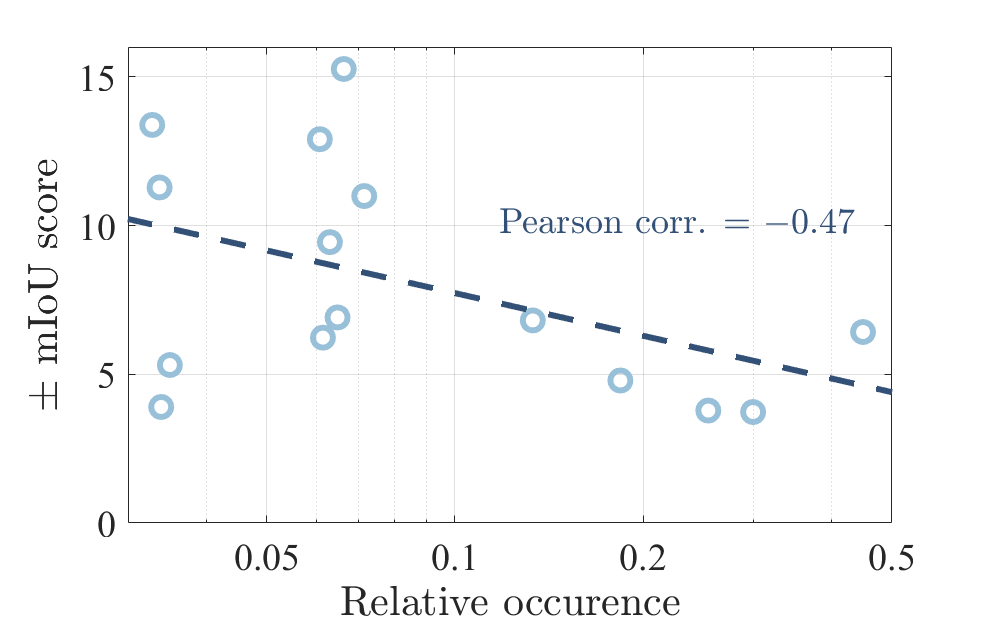}
    \caption{\textbf{Class imbalance.} We analyze the per-class IoU score for the results from~\cref{tab:sota_seg} in the main paper. Specifically, we contrast the relative occurrence of the $15$ foreground classes, with their absolute improvement in the mean IoU score on PSPNet~\cite{zhao2017pyramid}. The resulting negative correlation confirms that our approach implicitly mitigates class imbalances, since rare classes disproportionally benefit from additional generated samples. 
    }
    \label{fig:class_imbalance}
\end{figure}

We further provide an analysis of the impact of additional synthetic samples on rare classes, see~\cref{fig:class_imbalance}. For each of the $15$ foreground classes, we consider the absolute improvement of the mean IoU score on PSPNet~\cite{zhao2017pyramid}. This is contrasted with the relative class occurrence, defined as the fraction of images that contain any such instances. We observe a negative Pearson correlation coefficient of $-0.47$, which indicates that the generated samples help mitigate class imbalances. 

\section{Super-resolution discussion}\label{appendix:superres}

In~\cref{subsec:superresolution}, we devise a super-resolution approach that allows us to upsample generated images to a resolution of $256\times 256$. Specifically, we utilize a diffusion-based super-resolution model $\cG_\mathrm{SR}$ that takes generated images with a size of $128\times 128$ as a conditional input to the denoising U-Net. While it is conceivable to extend this approach to even higher resolutions $>256$, considered by some existing satellite segmentation baselines~\cite{li2021pointflow,zheng2020foreground}, we leave such investigations for future work due to the substantial computational demand of high-resolution diffusion models.

As a straightforward alternative to our super-resolution approach, we employ DDPM~\cite{ho2020denoising} to directly generate samples with a spatial size of $256\times 256$. The resulting accuracies of both approaches on the iSAID dataset are summarized in~\cref{tab:superres_exp}, considering two standard backbones FPN~\cite{kirillov2019panoptic} and SegFormer~\cite{xie2021segformer}. The experimental setting is analogous to~\cref{tab:sota_seg} in the main paper.

\begin{figure*}
    \centering
    \hspace{0.04\linewidth}
    \begin{overpic}
        [width=0.95\textwidth]{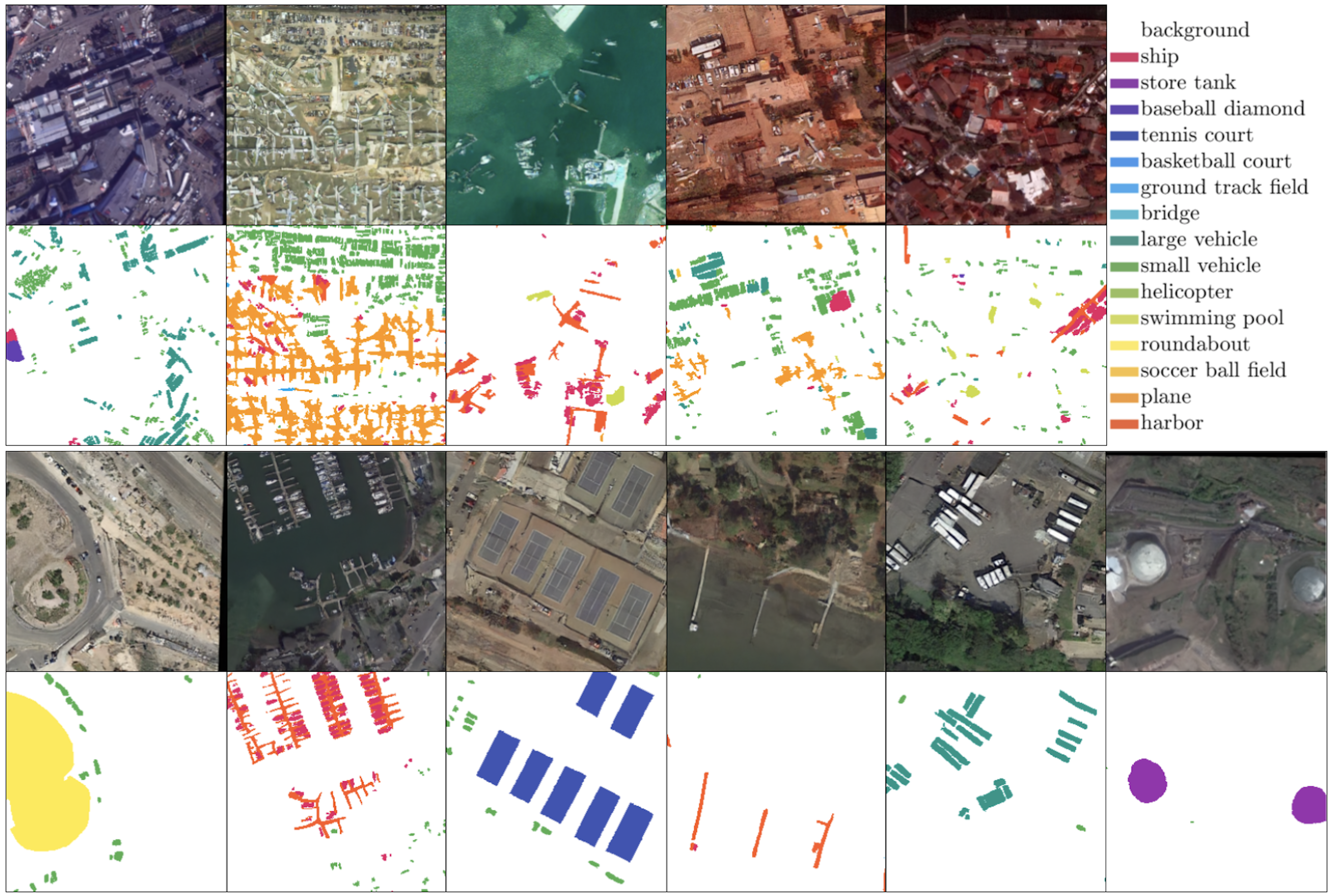}
        \put(-4,14){\rotatebox{90}{\large Ours}}
        \put(-4,45){\rotatebox{90}{\large DDPM-256}}
    \end{overpic}
    \caption{\textbf{Super-resolution comparison.} 
    We provide a qualitative comparison of image super-resolution to standard DDPM generations. 
    At resolutions $\geq 256$, DDPM exhibits unstable training behaviour, leading to severe artifacts -- both in terms of the saturation and contrast of obtained samples, as well as the overall semantic layout. In contrast, our super-resolution approach, outlined in~\cref{subsec:superresolution} of the paper, generates coherent and high-quality scenes (lower row).
    }
    \label{fig:superres_exp}
\end{figure*}

These results indicate that our super-resolution approach yields more consistent results compared to the direct DDPM generations. To investigate this effect, we additionally provide visualizations of the resulting pairs in~\cref{fig:superres_exp}. While both approaches yield comparable generations in terms of fine-scale details, the images and masks obtained with DDPM-256 are less coherent in the overall semantic layout. Moreover, the convergence behavior of DDPM-256 is less stable, producing erroneous image contrasts and saturations. While both approaches yield improvements for FPN~\cite{kirillov2019panoptic}, the lack of semantic coherence slightly decreases the performance of SegFormer~\cite{xie2021segformer}. Our super-resolution strategy effectively decouples the challenges of creating semantically consistent images through $\cG$ and recovering fine-scale details through $\cG_\mathrm{SR}$, leading to a superior downstream performance in~\cref{tab:superres_exp}.

\begin{table}
\begin{center}
\resizebox{0.82\linewidth}{!}{
\begin{tabular}{lcccc}
\toprule[0.2em]
&\multicolumn{2}{c}{\emph{FPN}} & \multicolumn{2}{c}{\emph{SegFormer}} \\ \cline{2-5}
    & IoU ($\uparrow$)& F1 ($\uparrow$) & IoU ($\uparrow$)& F1 ($\uparrow$)\\
    \toprule[0.2em]
    No synthetic  & 59.52 & 72.82& 60.95 & 74.18 \\
    DDPM-256   & 60.30 & 73.22 & 60.61 & 73.85 \\
    Ours & \textbf{60.65} & \textbf{73.69} & \textbf{62.13} & \textbf{75.10} \\

\bottomrule[0.1em]
\end{tabular}
}
\centering
    \caption{\textbf{Super-resolution.} 
     We compare our super-resolution approach to directly generating synthetic samples with DDPM, analogous to our approach in~\cref{subsec:approach}. We find that DDPM exhibits unstable training behaviour for resolutions $H=W\geq 256$ which results in a suboptimal downstream segmentation performance on iSAID. The obtained joint samples display notable artifacts, particularly in terms of the saturation and contrast of the generated images, refer to~\cref{fig:superres_exp} for a qualitative comparison.
     }
    \label{tab:superres_exp}
\end{center}

\end{table}

\section{Advanced data augmentation}
In~\cref{tab:gen_models_comp_b} of the main paper, we demonstrate consistent quantitative improvements in downstream segmentation tasks compared to existing generative approaches. 
Here, we evaluate the effectiveness of our approach against state-of-the-art augmentation techniques such as Cutout~\cite{devries2017improved}, CutMix~\cite{yun2019cutmix}, and Copy-Paste~\cite{ghiasi2021simple}. 

Cutout~\cite{devries2017improved} applies regional dropout in the input space for image classification. We adapt this to semantic segmentation by masking out random squares from both the image and its corresponding semantic mask. CutMix~\cite{yun2019cutmix} crops random regions from one image and pastes them onto another image, along with the corresponding masks. The instance segmentation augmentation approach Copy-Paste~\cite{ghiasi2021simple} copies connected semantic regions from one image to another. Compared to CutMix~\cite{yun2019cutmix}, such regions correspond to object instances instead of squares. 

We revisit the quantitative results from~\cref{fig:increasenumberofsynth} of the main paper, and report the resulting accuracies in~\cref{tab:advancedaug}. Specifically, we consider the iSAID dataset with a resampling ratio of $R=1$, and apply an FPN backbone. While all augmentation techniques enhance the performance, our approach yields the most significant quantitative improvements.

\begin{table}
\begin{center}
\resizebox{0.92\linewidth}{!}{
\begin{tabular}{c|cccc}
\toprule[0.2em]

    No additional samples & Ours &Cutout~\cite{devries2017improved} & CutMix~\cite{yun2019cutmix}&Copy-Paste~\cite{ghiasi2021simple} \\
    \toprule[0.2em]
      50.25& \textbf{51.11}& 50.47 & 50.60 & 50.51 \\

\bottomrule[0.1em]
\end{tabular}
}
\centering
    \caption{\textbf{Quantitative comparison of augmentation methods.} We compare our method to the recent augmentation techniques Cutout~\cite{devries2017improved}, CutMix~\cite{yun2019cutmix}, and Copy-Paste~\cite{ghiasi2021simple}. Across all experiments, we generate additional training pairs with a resampling ratio of $R=1$. The experimental setup is equivalent to the results on iSAID reported in~\cref{fig:increasenumberofsynth} of the main paper. 
     }
    \label{tab:advancedaug}
\end{center}

\end{table}
\section{Additional qualitative}\label{appendix:qualitative}
\begin{figure*}
    \centering
    \includegraphics[width=\linewidth]{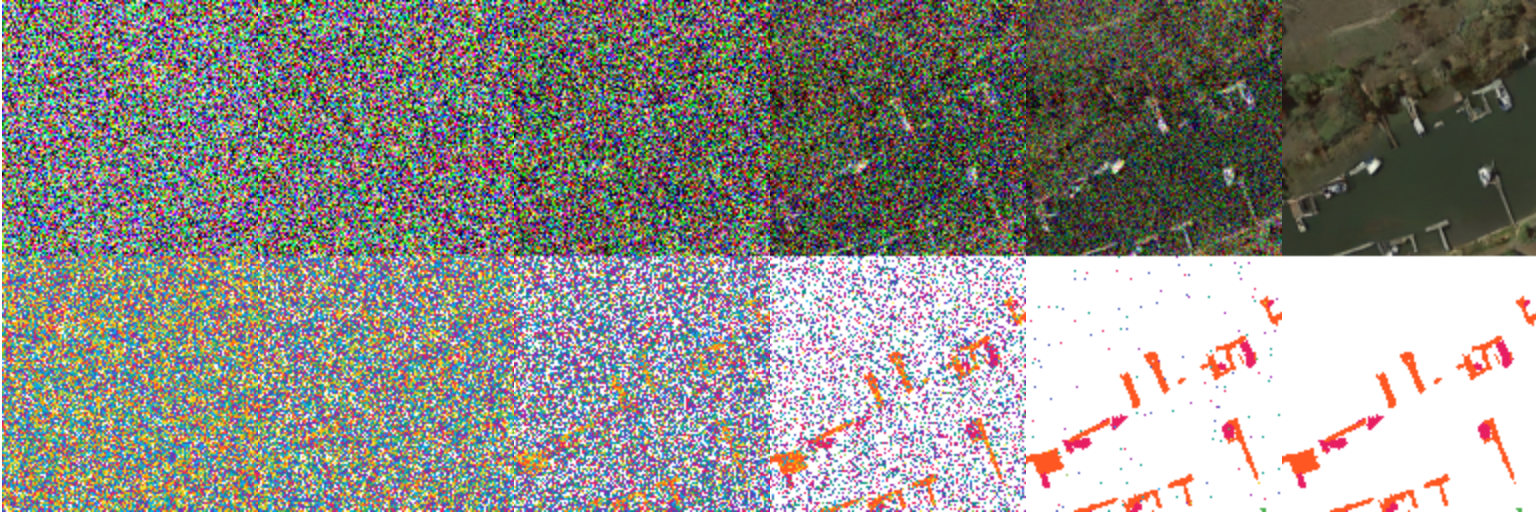}
    \caption{\textbf{Denoising process, qualitative.} We provide a qualitative example of the coupled denoising proposed in our approach. Similar to DDPM~\cite{ho2020denoising}, the novel training samples $(\mxi',\myi')$ emerge through an iterative scheme, reversing the forward Gaussian noising steps. 
    }
    \label{fig:denoising_visualize}
\end{figure*}

\begin{figure*}
    \centering
    \includegraphics[width=1.0\textwidth]{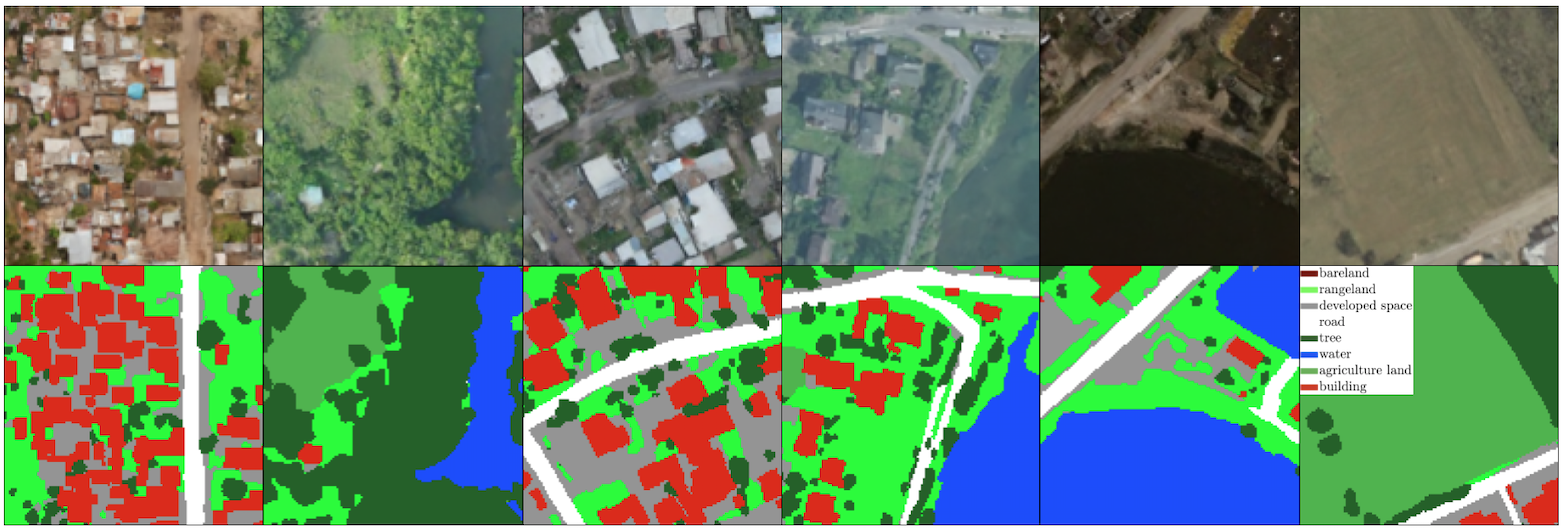}
    \caption{\textbf{Generated samples, OpenEarthMap~\cite{xia2023openearthmap}.} We display several generated joint instances $(\mxi',\myi')$ on OpenEarthMap~\cite{xia2023openearthmap}, obtained by the diffusion model $\cG$ detailed in~\cref{subsec:approach} of the main paper. 
    }
    \label{fig:syntheticvis_openearthmap}
\end{figure*}

For a complete picture, we provide several additional qualitative samples of different settings.
For once, we visualize the joint denoising process proposed in our approach in~\cref{fig:denoising_visualize}. We further show visualizations of generated samples on OpenEarthMap~\cite{xia2023openearthmap} in~\cref{fig:syntheticvis_openearthmap}, analogous to~\cref{fig:syntheticvis_isaid} for iSAID and~\cref{fig:syntheticvis_loveda} for LoveDA in the main paper. The semantic labels of OpenEarthMap are associated with land-cover classes, comparable to LoveDA. 

\begin{figure*}
    \centering
    \begin{overpic}
        [width=1.0\textwidth]{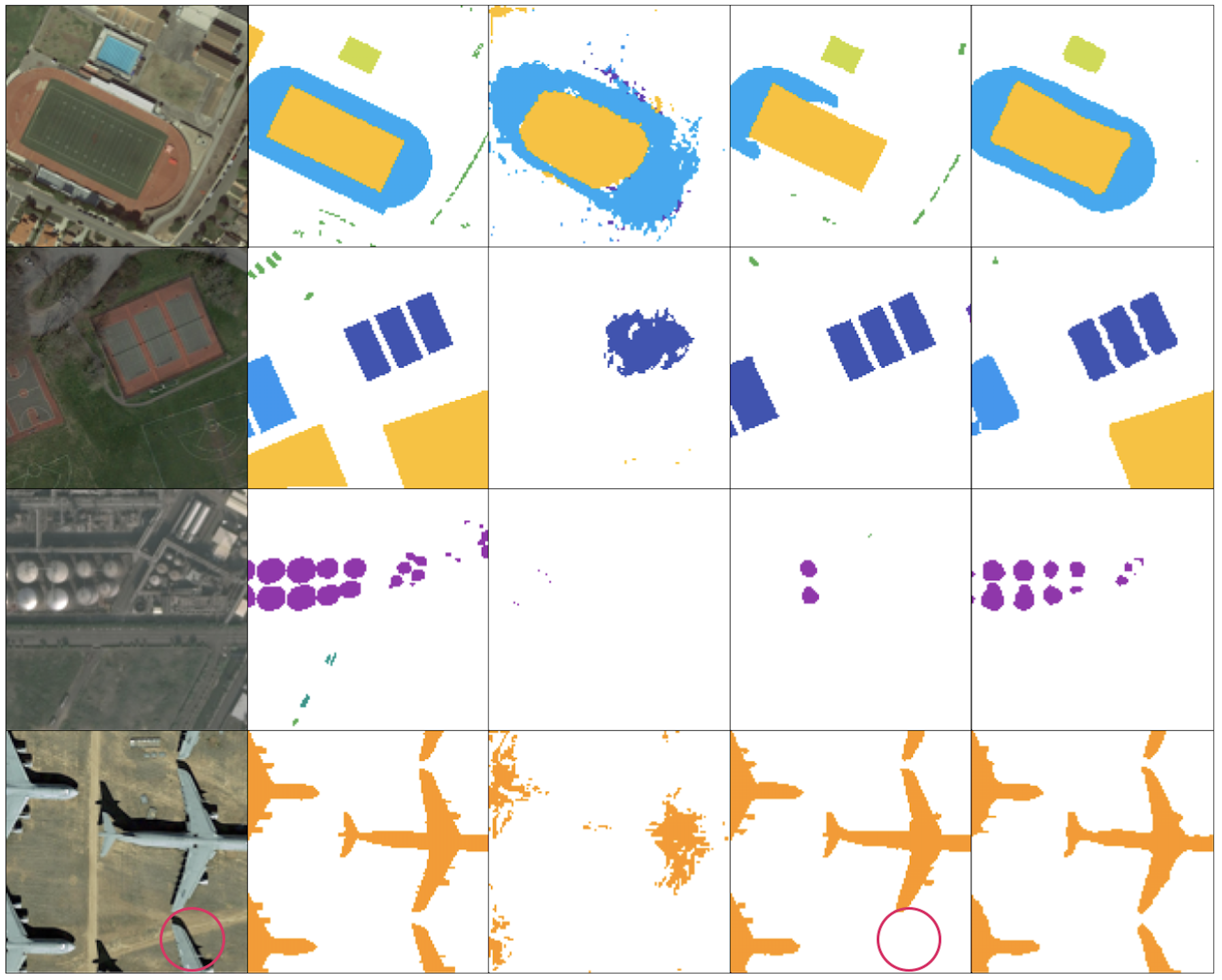}
        \put(2.75,83){\large Image $\mx$ (input)}
        \put(23.5,83){\large Mask $\my$ (g.t.)}
        \put(45,83){\large SemGAN}
        \put(66,83){\large SegDiff}
        \put(87,83){\large Ours}
    \end{overpic}
    \caption{\textbf{iSAID baseline comparison.} We contrast the semantic masks obtained with our approach to our two considered baselines~\cite{li2021semantic,amit2021segdiff}. These correspond to the results presented in~\cref{tab:gen_models_comp_b} in the main paper. SemGAN is primarily designed for conventional segmentation benchmarks such as CelebA~\cite{liu2018large}, whereas the generalization to imbalanced earth observation datasets is limited. Like ours, SegDiff yields high quality masks but individual regions are mislabeled more frequently (\eg red marker), as indicated by the quantitative results in~\cref{tab:gen_models_comp_b}.
    }
    \label{fig:segmentation_isaid}
\end{figure*}

\begin{figure*}
    \centering
    \begin{overpic}
        [width=1.0\textwidth]{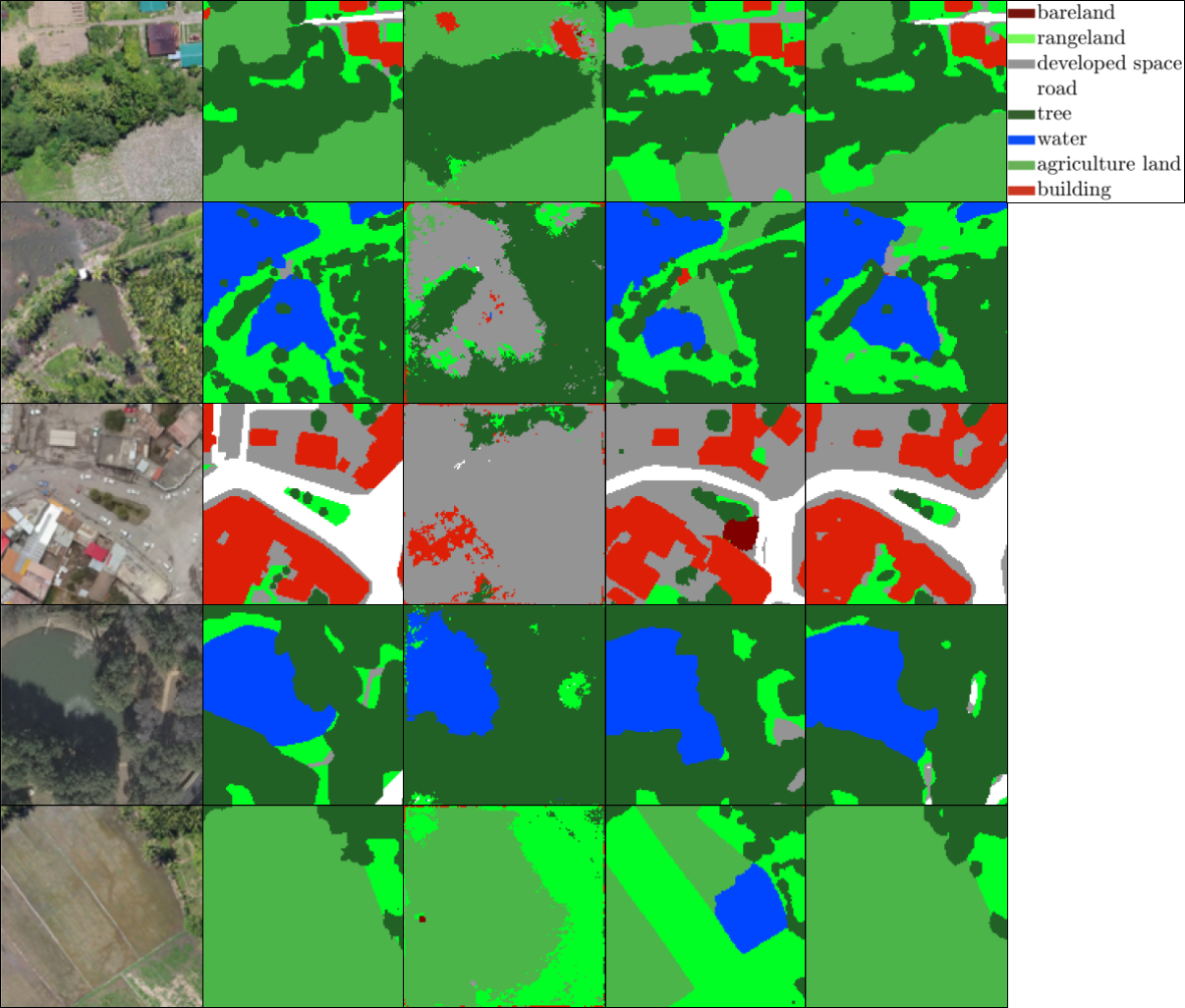}
        \put(1,89){\large Image $\mx$ (input)}
        \put(19.25,89){\large Mask $\my$ (g.t.)}
        \put(37.5,89){\large SemGAN}
        \put(55.5,89){\large SegDiff}
        \put(73.5,89){\large Ours}
    \end{overpic}
    \caption{\textbf{OpenEarthMap baseline comparison.} Analogously to~\cref{fig:segmentation_isaid}, we show a number of qualitative comparisons of our approach to our considered baselines~\cite{li2021semantic,amit2021segdiff} on OpenEarthMap.
    }
    \label{fig:segmentation_openearthmap}
\end{figure*}

In~\cref{fig:segmentation_isaid} and~\cref{fig:segmentation_openearthmap}, we visualize the predicted semantic masks for iSAID and OpenEarthMap, respectively. Compared to the two baselines SemGAN~\cite{li2021semantic} and SegDiff~\cite{amit2021segdiff}, our approach yields the most consistent results -- both in terms of accuracy and mask quality.

\begin{figure*}
    \centering
    \includegraphics[width=\textwidth]{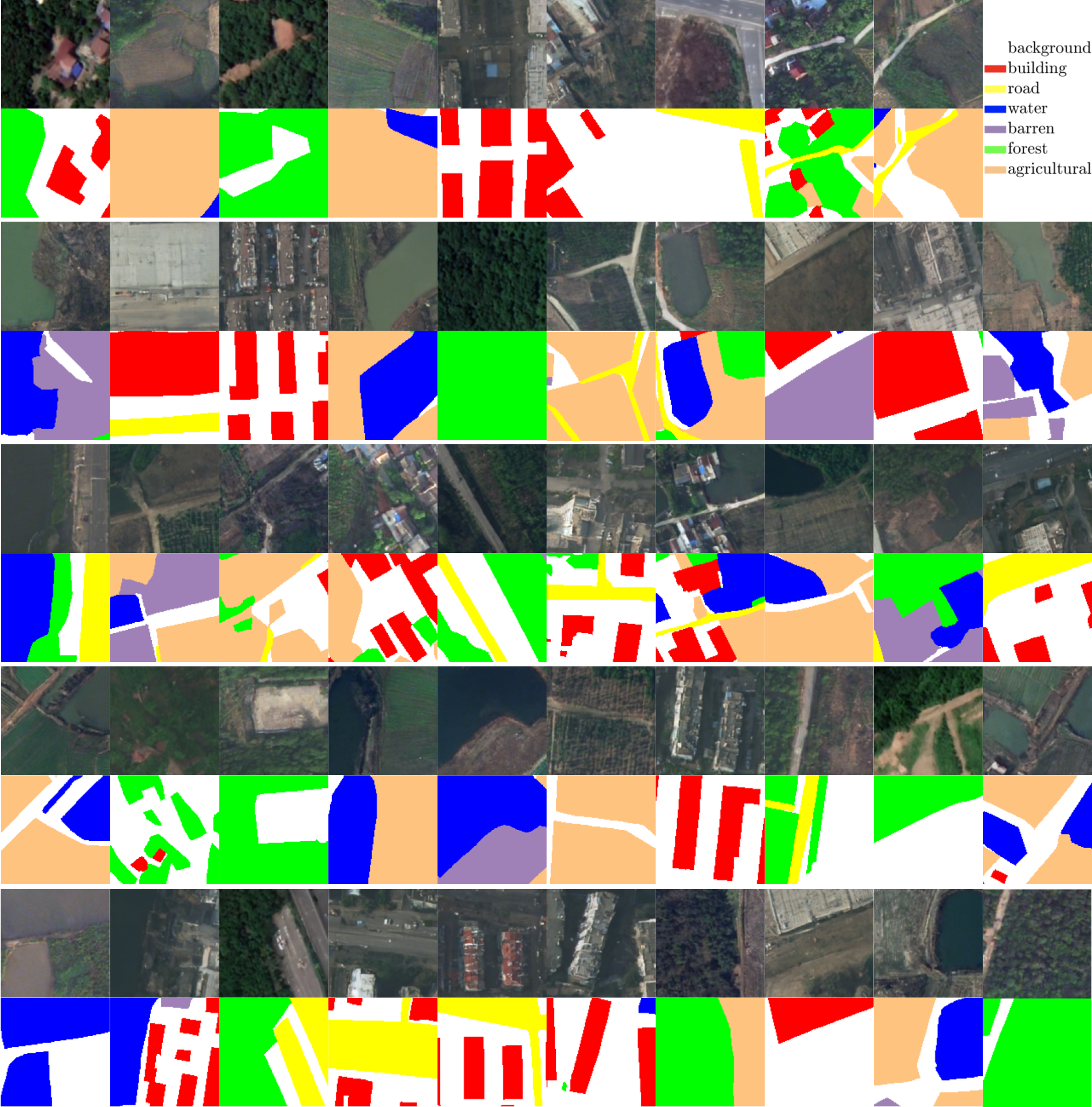}
    \caption{\textbf{LoveDA, qualitative.} We provide 49 random samples generated on LoveDA~\cite{wang2021loveda}, for an in-depth understanding of the quality of obtained samples. As usual, we show pairs of synthesized images $\mx$ and corresponding synthesized masks $\my$. 
    }
    \label{fig:loveda_many_samples}
\end{figure*}

Finally, we provide 49 random samples from LoveDA~\cite{wang2021loveda} in~\cref{fig:loveda_many_samples} for detailed insights into the obtained samples quality of our generative approach.

\begin{table*}
\begin{center}
\scalebox{0.65}{
\begin{tabular}{rcccccccccccccccccc}
\toprule[0.2em]
  & & \multicolumn{16}{c}{\emph{per class IoU} ($\uparrow$)}  \\  \cline{4-19}
  &  mIoU ($\uparrow$) & F1 ($\uparrow$) & BG & S & ST & BD & TC & BC &GTF & B&  LV & SV & HC & SP & R & SBF & P & H  \\
\bottomrule[0.1em]
\multicolumn{1}{l}{PFSegNet + $\cD$}  & 60.93  &  74.10 & 98.84  & 61.74 & 63.21&76.33	&84.25 &	48.98 &  54.13 &34.95 & 61.66&  41.60& 26.52 & \textbf{50.43} & 67.95 &66.11  & 81.41& 56.80 \\
\multicolumn{1}{l}{PFSegNet + $\cD\cup\cD'$}& \textbf{63.71} & \textbf{76.37} & \textbf{98.93} & \textbf{63.88}  & \textbf{67.58} & \textbf{77.11} &  \textbf{87.70} & \textbf{57.96} & \textbf{58.01} & \textbf{40.22} & \textbf{62.91} &\textbf{44.70}  & \textbf{27.63} & 50.42 & \textbf{70.28} &  \textbf{69.59} & \textbf{82.75} & \textbf{59.71} \\
 \hline
\multicolumn{1}{l}{FarSeg + $\cD$}   & 62.28  & 75.16 & 98.84 &\textbf{62.45}&	\textbf{68.63}	&\textbf{76.76}&	86.15&	57.14&	54.39&	\textbf{38.95}&	\textbf{61.49}&	\textbf{41.08}&	27.50&	\textbf{45.55}&	71.53&	70.45&	81.09&	54.53 \\
\multicolumn{1}{l}{FarSeg + $\cD\cup\cD'$} & \textbf{62.95}& \textbf{75.72} & \textbf{98.87} &62.33   &  68.16& 74.83 &\textbf{86.60} & \textbf{57.73}&	\textbf{57.22}&	38.56&	61.35& 40.20&	\textbf{30.10}&	45.53&	\textbf{74.23}&	\textbf{72.69}&	\textbf{81.69}&	\textbf{57.09} \\
 \hline
\multicolumn{1}{l}{SegFormer  + $\cD$}   & 60.95 & 74.18 & 98.83 & 61.91 & 63.58 & \textbf{74.35} & 84.72 & 54.01 & 57.74 & \textbf{40.37} & 58.20 & 34.32 & \textbf{32.48} & 42.27 & 68.25 & 72.67 & 78.53 & 52.83  \\
\multicolumn{1}{l}{SegFormer + $\cD\cup\cD'$ }   & \textbf{62.13}&\textbf{75.10} & \textbf{98.88} & \textbf{64.17} & \textbf{64.86} & 74.22 & \textbf{85.86} & \textbf{58.76} & \textbf{58.01} & 40.19 & \textbf{59.53} & \textbf{35.93} & 29.69 & \textbf{46.20} & \textbf{69.92} & \textbf{72.73} & \textbf{79.89} & \textbf{55.21} \\
\hline
\multicolumn{1}{l}{FPN + $\cD$}   & 59.52&  72.82 &98.78&	58.66&	63.72&	76.39&	84.97&	55.32&	58.05&	36.15&	56.82&	34.40&	\textbf{23.82}&	44.52&	63.60&	70.45&	76.57&	50.14\\
\multicolumn{1}{l}{FPN + $\cD\cup\cD'$} &\textbf{60.65} &\textbf{73.69}& \textbf{98.82}&	\textbf{59.17}&	\textbf{64.87}&	\textbf{76.53}&	\textbf{85.97}&	\textbf{55.35}&	\textbf{58.28}&	\textbf{36.71}&	\textbf{57.96}&	\textbf{34.54}&	23.64&	\textbf{46.66}&	\textbf{69.79}&	\textbf{71.34}&	\textbf{77.70}&	\textbf{53.05} \\
\hline
\multicolumn{1}{l}{PSPNet + $\cD$}  & 48.95&63.13  &98.35&	46.33&	46.43&	68.27&	77.92&	38.63&	46.76&	21.45&	48.28&	18.03&	22.50&	35.96&	55.80&	55.24&	66.36&	36.88  \\
\multicolumn{1}{l}{PSPNet + $\cD\cup\cD'$}& \textbf{56.54}& \textbf{70.16} & \textbf{98.60}&	\textbf{51.12}&	\textbf{59.33}&	\textbf{73.58}&	\textbf{84.15}&	\textbf{52.01}&	\textbf{56.20}&	\textbf{32.44}&	\textbf{52.01}&	\textbf{24.45}&	\textbf{26.40}&	\textbf{42.87}&	\textbf{67.08}&	\textbf{70.50}&	\textbf{70.14}&	\textbf{43.69}\\
\bottomrule[0.1em]
\end{tabular}
}
\centering
    \caption{\textbf{Per-class segmentation scores on iSAID $256\times256$.}
    We provide a detailed analysis of the per-class segmentation scores on iSAID. Specifically, we report mean IoU scores for approaches tailored for high-resolution satellite imagery~\cite{li2021pointflow, zheng2020foreground} and the general-purpose segmentation models SegFormer~\cite{xie2021segformer}, FPN~\cite{kirillov2019panoptic}, and PSPNet~\cite{zhao2017pyramid}. Each model is trained on the combined dataset of original and generated samples $\cD\cup\cD'$, and compared against models trained solely on the original data $\cD$. For a majority of classes, the synthesized data yields marked improvements in performance. We abbreviate the 16 semantic classes with the following acronyms: background (BG), ship (S), store tank (ST), baseball diamond (BD), tennis court (TC), basketball court (BC), ground track field (GTF), bridge (B), large vehicle (LV), small vehicle (SV), helicopter (HC), swimming pool (SP), roundabout (R), soccer ball field (SBF), plane (P), harbour (H).}
    \label{tab:allclasses_isaid}
\end{center}

\end{table*}

\end{document}